\PassOptionsToPackage{numbers,sort&compress}{natbib}
\documentclass[anon]{hld2025}

\usepackage[utf8]{inputenc}
\usepackage{amsmath, amssymb, amsfonts}
\usepackage{graphicx}
\usepackage{booktabs} 

\usepackage{tikz}
\usepackage{wrapfig}
\usepackage{float} 
\usepackage[font=small,skip=5pt]{caption} 

\usepackage{hyperref}

\title{Adapting to High Dimensional Concept Space with Metalearning}

\author{
\begin{center}
    {\large \bf Max Gupta}$^1$ \\
    \texttt{mg7411@princeton.edu} \\
    $^1$Department of Computer Science, Princeton University
\end{center}
}

\date{July 11, 2025}

\begin{document}
\maketitle

\begin{abstract}
Rapidly learning abstract concepts from limited examples is a hallmark of human intelligence. This work investigates whether gradient-based meta-learning can equip neural networks with inductive biases for efficient few-shot acquisition of discrete concepts. I compare meta-learning methods against a supervised learning baseline on Boolean concepts (logical statements) generated by a probabilistic context-free grammar (PCFG). By systematically varying concept dimensionality (number of features) and recursive compositionality (depth of grammar recursion), I delineate between complexity regimes in which meta-learning robustly improves few-shot concept learning and regimes in which it does not. Meta-learners are much better able to handle compositional complexity than featural complexity. I highlight some reasons for this with a loss landscape analysis demonstrating how featural complexity increases the roughness of loss trajectories, allowing curvature-aware optimization to be more effective than first-order methods. I find improvements in out-of-distribution generalization on complex concepts by increasing the number of adaptation steps in meta-SGD, where adaptation acts as a way of encouraging exploration of rougher loss basins. Overall, this work highlights the intricacies of learning compositional versus featural complexity in high dimensional concept spaces and provides a road to understanding the role of 2nd order methods and extended gradient adaptation in meta-learning. \end{abstract}

\section{Introduction}

Human learners are able to learn concepts from remarkably little data \cite{Lake2015bpl}. Early models of concept learning that were able to match human performance leveraged Bayesian inference and symbolic modeling \cite{Goodman2008lot, TenenbaumGriffiths2001, Feldman2000concept}, explaining human performance as an approximation of rational inference. However, these methods often suffered from computational intractability. Recent approaches have combined Bayesian modeling with neural networks through metalearning, allowing models to match Bayesian performance with a fraction of the data \cite{LakeBaroni2023}, thus providing a more plausible mechanism to explain the remarkable sample efficiency of human concept learning. Few approaches have tested the opposite end of the spectrum and explored entirely gradient-based meta-learning methods and effects on sample efficiency and generalization in the concept learning domain. Furthermore, while meta-learning has achieved impressive results across domains including perception, control, and reasoning, existing evaluations often focus on performance within fixed datasets, leaving underexplored how meta-learning behaves as task complexity systematically increases to higher dimensionalities. 

In this work, I study the limits of end-to-end, gradient-based meta-learning in a \emph{Boolean concept learning task}, in a setting that allows precise control over compositional and featural complexity via a probabilistic context free grammar (PCFG). By extending a PCFG-based concept generator developed initially by Goodman et al. 2008 \cite{Goodman2008lot}, I independently vary \textit{featural dimensionality} (number of binary input features) and \textit{compositional depth} (logical recursion) to create a distribution of tasks with increasing structural complexity. This setting enables a study of how out-of-distribution (OOD) generalization from meta-learning scales with the complexity of the underlying concept space. 

I compare gradient-based meta-learning (Meta-SGD) against standard supervised learning (SGD) on few-shot Boolean classification tasks. Results show that meta-learning is incredibly robust at handling increased compositional recursions, in line with previous work \cite{conklin-etal-2021-meta, LakeBaroni2023}, but suffers with increased featural dimensionality. To explain this, I show how increasing dimensionality results in an increase in the roughness of the loss landscape of the 'concept basin'. As a potential remedy, I find empirical evidence that increasing the number of adaptation steps can reliably help a meta-learner navigate these rougher loss landscapes, to find generalization-friendly weight initializations. I also present early evidence of how curvature awareness (the second-order gradient term) helps meta-learning effectively navigate different concept complexities by analyzing the Hessian trace of increasingly difficult concept landscapes. An analysis of loss landscape roughness reveals a strong correlation between landscape curvature and relative gains from meta-learning, partially explaining when and why meta-learning is effective in high dimensional concept learning.

\section{Related Work}\label{sec:related}

\textbf{Gradient-based meta-learning.}  
MAML \citep{Finn2017maml} introduced a framework for learning model initializations that adapt quickly via gradient descent. Meta-SGD \citep{Li2017metasgd} extends this by learning per-parameter step sizes, enabling one-step adaptation. First-order approximations such as FOMAML and Reptile \citep{Nichol2018firstorder} omit Hessian terms to reduce cost, yet their performance often matches full MAML on vision tasks. Theoretical analyses highlight that second-order updates embed an implicit contrastive objective, which can improve generalization on harder tasks \citep{Kao2022contrastive}. 

\textbf{Compositional generalization and concept learning.}  
Symbolic rule induction methods, such as Bayesian Program Learning (BPL) \citep{Lake2015bpl} and the Rational Rules model \citep{Goodman2008lot}, achieve human-level one-shot learning by leveraging explicit grammars. However, they require handcrafted generative models and search. Neural sequence-to-sequence models struggle with systematic generalization on tasks like SCAN \citep{Lake2018scan}, and neural meta-learners underperform on benchmarks like CURI \citep{Vedantam2023curi}. Meta-learning has recently been used to improve compositional generalization in NLP \citep{conklin-etal-2021-meta} and neuro-symbolic reasoning systems \citep{Ye2022nemesys}, but its role in Boolean concept induction remains underexplored. A theoretical framework for compositional generalization in neural networks was recently proposed \citep{Arora2024theoretical}, and surveys highlight the challenges and opportunities for compositional AI \citep{Shen2024survey}. Few studies have explicitly engaged in an exploration of how increasingly recursive structure affects the ability of a meta-learner. In this work, I take a first step in this direction in a discrete, Boolean setting to isolate logical complexity as a predictive variable for metalearning performance.
\section{Experimental Setup}\label{sec:methods}

The experimental setup starts by modifying the concept-generating PCFG from Goodman et al. 2008 \citep{Goodman2008lot} to explicitly control compositionality (recursion depth $D \in \{3, 5, 7\}$) and feature dimensionality (the number of literals $F \in \{8, 16, 32\}$). The grammar’s production rules and their sampling probabilities are given by :

\begin{quote}
\begin{tabbing}
\hspace{1cm}\=\hspace{1cm}\=\hspace{1.5cm}\=\kill

C   \> $\rightarrow$ \> L                \> p = 0.30 \\
C    \> $\rightarrow$ \> $\neg C$          \> p = 0.20 \\
C    \> $\rightarrow$ \> (C $\wedge$ C)    \> p = 0.25 \\
C    \> $\rightarrow$ \> (C $\vee$ C)      \> p = 0.25 \\[0.5\normalbaselineskip]

L   \> $\rightarrow$ \> $x_i$, where $x_i \in \mathcal{X} = \{x_1,\dots,x_F\}$
\end{tabbing}
\end{quote}

For each concept $C$, I generate a $K_{\text{shot}}$-sized support set $S_C$ (with $K_{\text{shot}} = 5$ positive and 5 negative labeled examples $(\mathbf{x}, C(\mathbf{x}))$), and a query set $Q_C$, both sampled from the Boolean input space $\{0,1\}^F$.

Each meta-learning episode samples a concept $C\sim\text{PCFG}(F,D)$ and creates support/query sets $S_C, Q_C$ from $\{0,1\}^F$ ($K_{\text{shot}}=10$, $K_{\text{qry}}=20$). Inner-loop adaptation performs $K_{\mathrm{adapt}}$ gradient updates: $\theta^{(k+1)} =\theta^{(k)}-\alpha\odot \nabla_{\theta^{(k)}}\mathcal{L}_{S_C}(\theta^{(k)})$, yielding $\theta_{\mathrm{adapt}}$. The outer-loop updates $\mathcal{L}_{\text{meta}}(C)=\mathcal{L}_{Q_C}(\theta_{\mathrm{adapt}})$ and back-propagates through the inner loop to update $(\theta_{\mathrm{init}},\alpha)$ with the Adam optimizer \cite{Kingma2014adam}.

Episodes contain both K-shot training examples ($S_C$) and held-out evaluation examples ($Q_C$), ensuring meta-learners are rewarded only for configurations that generalize within tasks. This systematic complexity manipulation enables controlled study of how logical structure affects meta-learning performance. 

All methods use a 5-layer MLP (128 hidden units/layer, ReLU, sigmoid output). I compare  models trained with three stochastic gradient descent (SGD) learning algorithms, varying the order of the gradients and adaptation steps: 1st-Order and 2nd-Order Meta-SGD with both 1 and 10 adaptation (gradient) steps and regular SGD: training from scratch per task using Adam (learning rate 0.001) on $S_C$.

\begin{figure}[H]
    \centering
    \includegraphics[width=1.0\textwidth]{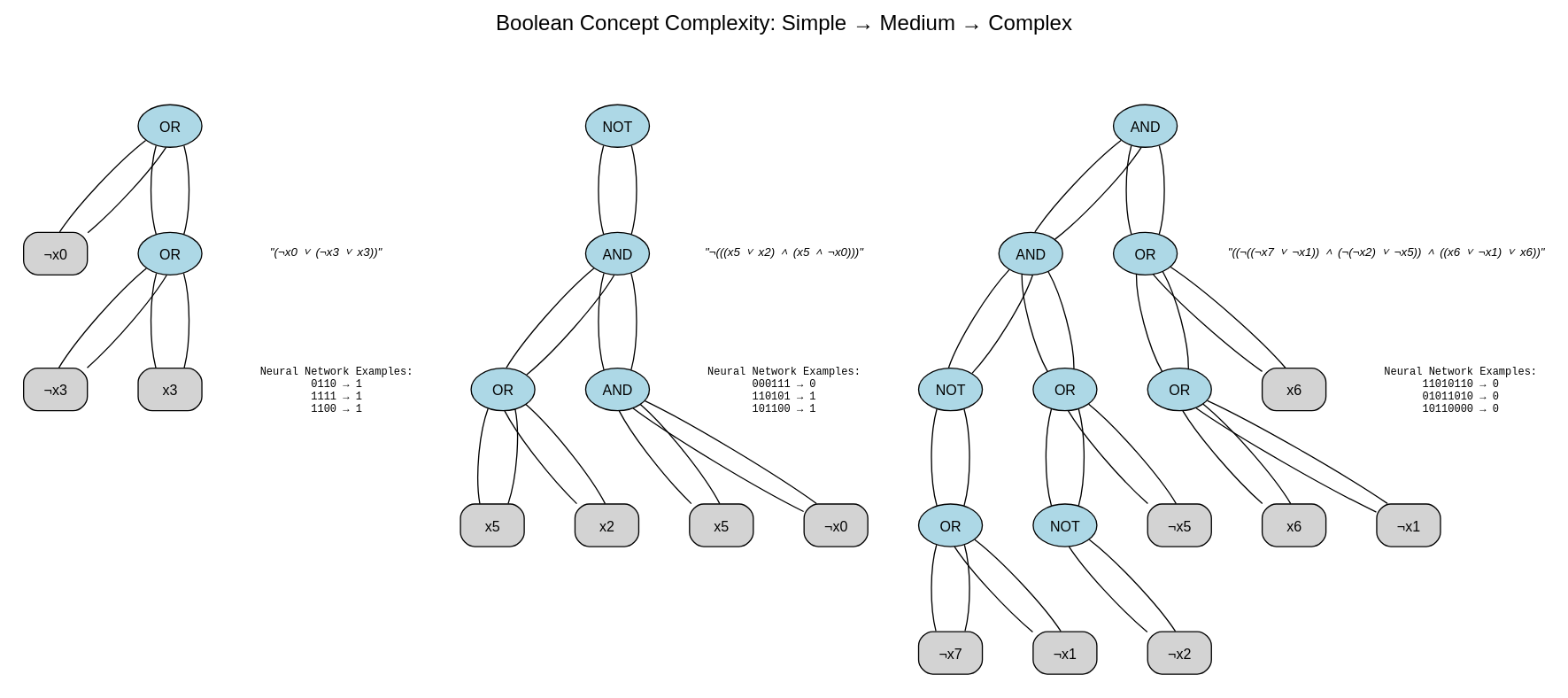}
    \caption{The PCFG parse trees of concepts with increasing complexity. Here compositional depth is visualized as the depth of the parse tree on the vertical axis, feature dimensionality is visualized as the width of the parse tree on the horizontal axis. Examples show how PCFG-generated concepts scale from simple to complex logical structures. \textbf{Left}: Simple concept with 2 features and depth 3. \textbf{Center}: Medium complexity with 3 features and depth 4. \textbf{Right}: Complex concept with 5 features and depth 5. Neural networks see only the bit-string input of features and ideally learn to infer the logical structure of the underlying concept over successive trials.}
    \label{fig:pcfg}
\end{figure}

Increasing $K_{adapt}$ allows more extensive search in the task-specific loss landscape, incrementally adjusting the MLP's decision boundaries to correctly classify support set examples. Meta-SGD models were meta-trained for 10,000 episodes. All evaluations were averaged over 5 random seeds on 1,000 unseen tasks (like those shown in Figure 1). For trajectory comparisons, SGD is trained for steps equivalent to processing a fixed total number of samples. Performance is assessed using final mean accuracy and data efficiency (samples required to reach 60\% accuracy, Appendix~\ref{app:data_efficiency}).

\section{Results}\label{sec:results}
\begin{figure}[H]
    \centering
    \includegraphics[width=0.95\linewidth]{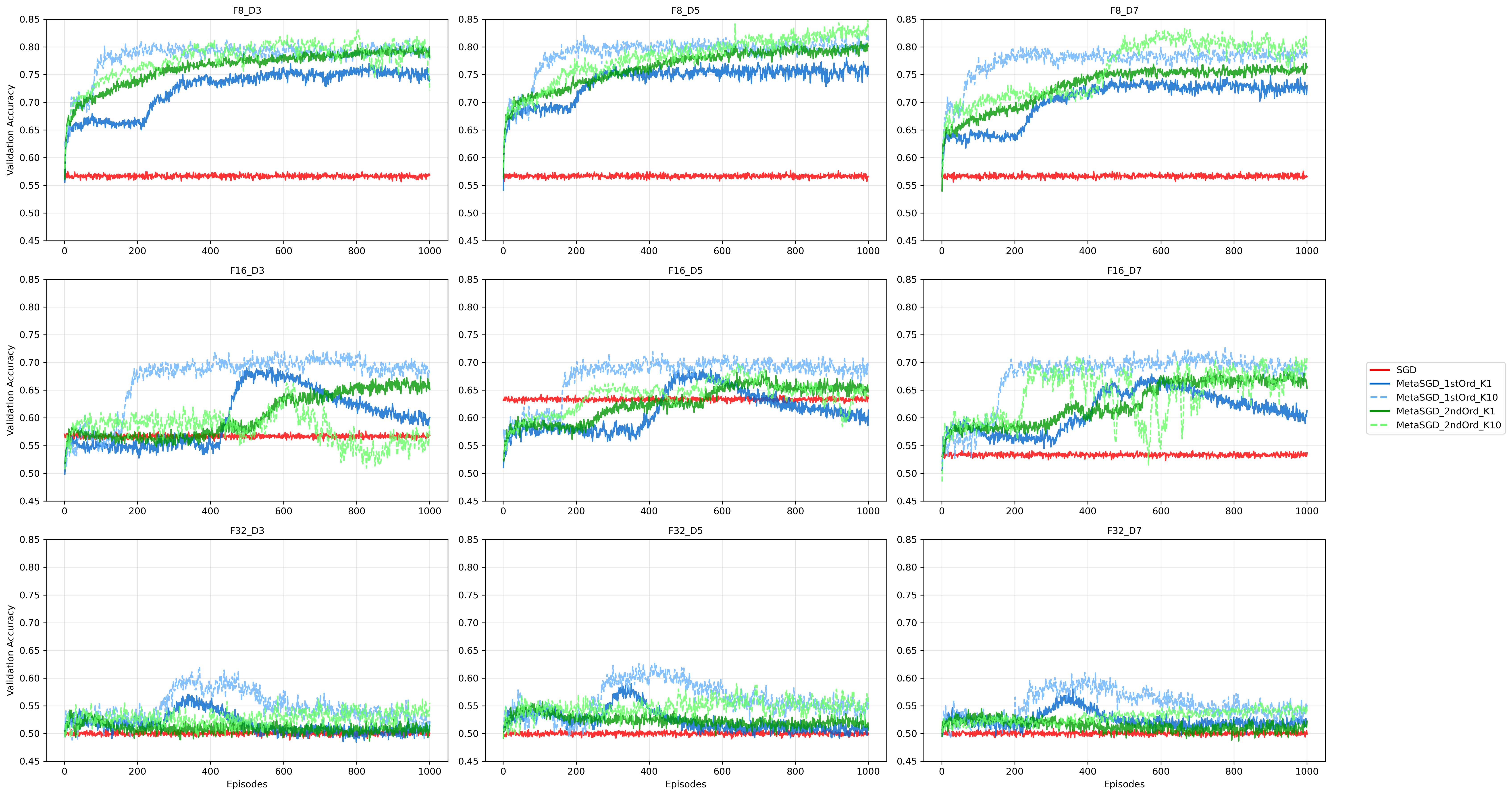} 
    \caption{First order meta-SGD (blue lines) versus second order meta-SGD (green lines) and vanilla SGD (red line) performance over increasingly complex Boolean concepts. Featural complexity (number of literals) increases along  (rows) and concept depths (columns) over normalized training episodes. }
    \label{fig:accuracy_trajectories}
\end{figure}

Figure~\ref{fig:accuracy_trajectories} shows learning trajectories across a sweep of feature dimensionalities ($F$) and concept depths ($D$), averaged for noise over 5 seeds. Meta-SGD methods demonstrate clear advantages over SGD, learning faster and converging to higher accuracies, particularly for $F=8$ and $F=16$. First-order meta-SGD with increased adaptation steps (K=10) matches or exceeds second-order performance.

Meta-learning demonstrates substantial data efficiency advantages (Appendix~\ref{app:data_efficiency}), with 1st-order Meta-SGD using K=10 adaptation steps requiring orders of magnitude fewer samples than SGD to reach 60\% accuracy, particularly at $F=8, D=3$.

At $F=32$, all methods show significant performance drops as featural complexity peaks (bottom row). While second-order methods outperform in low-complexity settings (top row, $F=8$), first order methods with added gradient steps outperform in the higher featural dimension settings (middle and bottom rows), suggesting that curvature-awareness is relatively advantageous for simpler concepts only (as we will see in the next section, simpler concepts encode smoother loss landscapes). Even in high-dimensional regimes, increased adaptation (K=10) yields the largest relative improvements, suggesting extensive adaptation becomes crucial when feature space grows. 

\section{Loss Landscape Analysis}\label{sec:landscape}

Behaviorally, we find that meta-learning is highly effective across concept depth, but offers rapidly diminishing returns as we scale the number of features per concept, correpsonding to the depth and breadth of the PCFG in Figure 1.  In this section, I take a closer look at how the learning process diverges between meta-SGD and SGD over the concept classes tested above by visualizing the loss landscapes during the training process itself.    

\subsection{Methodology}\label{sec:roughness_method}

I define \textit{roughness} as optimization instability: trajectory variation during training. Equation (1) defines a metric of roughness that extracts loss sequences $L = [l_1, l_2, \ldots, l_T]$, normalized over 200 episodes of training, applies Gaussian smoothing ($\sigma=1$), computes discrete second derivatives $\nabla^2 L_i = l_{i+1} - 2l_i + l_{i-1}$, and calculates:

\begin{equation}
\text{Roughness} = \frac{\text{std}(\nabla^2 L)}{\text{mean}(|\nabla^2 L|) + \epsilon}
\end{equation}

where $\epsilon = 10^{-8}$ prevents division by zero. This normalized measure captures optimization instability, with higher values indicating more erratic training behavior characteristic of rugged loss landscapes, and lower values representing smoother convergence on more navigable terrain.

The goal is to measure how introducing meta-learning (2nd order gradients) changes the loss function across increasingly complex concepts, each define by their own loss landscapes (Figure~\ref{fig:landscape_navigation}). 

\subsection{Complexity-Dependent Landscape Topology}

\begin{figure}[ht]
    \centering
    \includegraphics[width=1.0\textwidth]{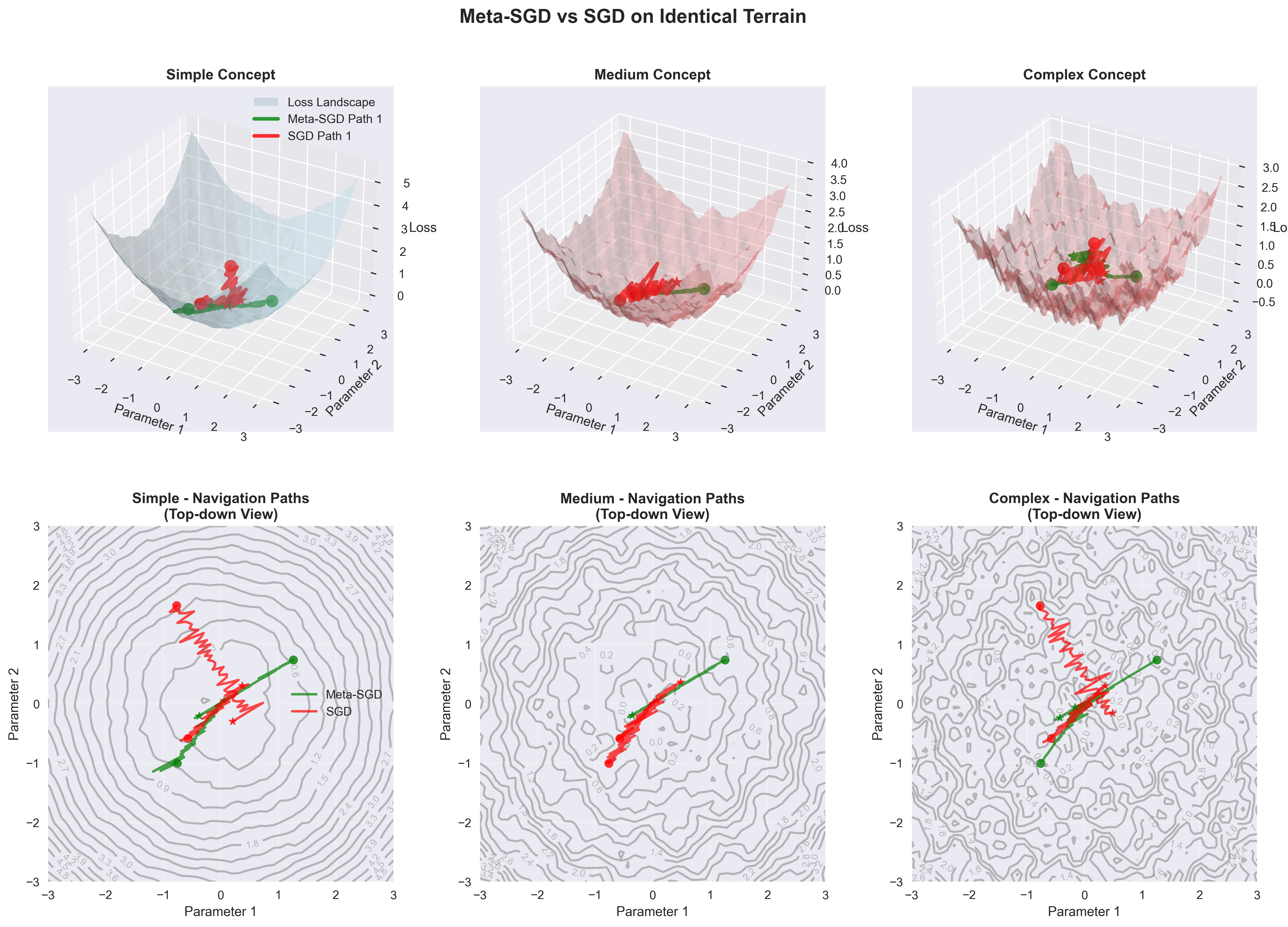}
    \caption{Meta-SGD and SGD operate on the same concept loss landscapes (determined by task structure and architecture), but meta-learning learns more efficient navigation strategies (shorter paths to the solution point - the bottom-most point in each loss 'basin'). \textbf{Top row}: 2D loss landscapes for simple, medium, and complex Boolean concepts show identical topology regardless of optimization method. \textbf{Middle row}: 3D visualizations reveal the terrain both algorithms must navigate, with complexity-dependent roughness. \textit{Note: due to computational intractability, loss landscapes are local approximations.}}
    \label{fig:landscape_navigation}
\end{figure}

I analyzed loss landscapes by sampling random directions in parameter space to approximately visualize the optimization challenge posed by each class of boolean concept difficulty. Boolean concept complexity  increases roguhness of landscape topology, creating different optimization challenges (Figure~\ref{fig:landscape_navigation}).

A quantitative analysis reveals systematic patterns: simple concepts (2-3 literals) exhibit smooth, quasi-convex landscapes with few local minima (on average 0.3±0.1, roughness = 0.0002); medium concepts (4-6 literals) show moderately rugged topology (1.2±0.4 minima, ~4x roughness increase); complex concepts (7+ literals) display highly rugged landscapes with multiple local minima (2.8±0.6, ~12x roughness increase). In other words, featural complexity creates characteristic landscape patterns, which map to optimization difficulties - the number and distribution of local minima. This partially explains the failures of meta-SGD to learn featurally dense concepts ( 

\subsection{Meta-SGD Learns Shorter Optimization Paths}

Our analysis reveals that meta-learning learns efficient navigation strategies for rugged landscapes comparared to vanilla SGD. 

As documented in Appendix A, Meta-SGD achieves a 90-99\% reduction in trajectory length (the geodesic length) compared to SGD baseline across all concept complexities. This dramatic improvement in navigation efficiency directly translates to performance improvements: +15.5\% for simple concepts, +34.1\% for medium concepts, and +11.1\% for complex concepts.

The largest improvement occurs at medium complexity, where Meta-SGD balances exploration and convergence. Meta-SGD consistently produces trajectories with lower variance in second derivatives, indicating more stable convergence that translates to better performance. Detailed curvature and Hessian trace analysis (see Appendix~\ref{app:landscape_curvature}) confirms that meta-learning learns more efficient pathways through identical loss surfaces, establishing a quantitative framework for predicting meta-learning utility from landscape properties.

Relative use of additional adaptation steps (K=10 vs K=1) scales with concept complexity (see Appendix~\ref{app:k_comparison}). Simple concepts show modest 5-8\% improvement from K=10 steps, while complex concepts demonstrate 15-20\% gains, supporting the intuitive argument that rugged landscapes require multiple steps to escape local minima.

\section{Discussion}\label{sec:discussion}

Our systematic manipulation of two orthogonal complexity dimensions—featural dimensionality ($F \in \{8, 16, 32\}$) and compositional depth ($D \in \{3, 5, 7\}$)—reveals fundamentally different challenges for meta-learning in Boolean concept acquisition. This controlled experimental design illuminates how different aspects of problem structure interact with optimization landscapes and meta-learning effectiveness.

\textbf{Compositional Complexity} Across all experiments, meta-learning demonstrates remarkable robustness to increasing compositional depth. Moving from $D=3$ to $D=7$ (simple to deeply nested logical structures) shows minimal performance degradation for Meta-SGD, while SGD suffers substantially. Our loss landscape analysis reveals why: compositional complexity primarily affects the logical structure within concept space but preserves relatively navigable optimization surfaces. The PCFG's recursive depth creates more intricate Boolean relationships without altering the smoothness of parameter space traversal. Meta-learning's learned initialization and adaptive step sizes prove particularly effective at discovering these hierarchical patterns within reasonable adaptation budgets.

\textbf{Featural Complexity} In stark contrast, increasing featural dimensionality poses severe challenges for all methods, with performance collapsing dramatically at $F=32$. This reveals a deeper truth about the nature of concept learning: while logical complexity (compositionality) can be handled through better optimization strategies, dimensional complexity  alters the search space. The explosion from $2^8$ to $2^{32}$ possible input configurations under high data sparsity creates loss landscapes so rugged and high-dimensional that meta-learning alone cannot overcome the curse of dimensionality. Our roughness analysis confirms that featural complexity creates exponentially more challenging optimization terrain than compositional complexity and provides insight for future concept learning work.

\textbf{Landscape Implications.} This dual-axis analysis reveals that not all forms of "complexity" are equivalent from an optimization perspective. Compositional depth affects the logical relationships that must be learned but preserves loss surface properties. The extent to which this is true in higher dimensional setting and with more complex models deserves further investigation. Furthermore, these loss trajectories were randomly sampled in a small, locally convex area of the total loss landscape, but further probing with more compute could reveal more complex topologies than the approximations provided. 

These findings suggest that meta-learning is particularly well-suited for domains where complexity arises from structural relationships rather than raw dimensionality, explaining its success in few-shot learning across compositionally rich but feature-constrained domains \citep{Nichol2018firstorder,Kao2022contrastive}. 


\section{Future Work}

An unanswered and intriguing question in the study of compositionality is how compositional concepts are embedded in the high-dimensional vector space parametrized by the hidden units of a sufficiently deep neural network. In this case, we have already seen that simple MLP's can accurately classify boolean concepts of low to medium compositional depth. I would be very curious to examine the learned representations of the MLP's to explore if any regular structure emerges. Given that the concept space itself is regular and well-defined hierarchically, I would suggest a simple Principal Component Analysis (PCA) on the hidden units of the MLP's trained in the above experiments. In a separate vein of inquiry, the extent to which I tested out of distrbution grammars was very limited - concept length generalization to out-of-distribution grammars would allow us to test the effectiveness and range of the inductive bias endowed by metalearning.  

\section{Conclusion}

This investigation across featural dimensionality and compositional depth reveals when and why meta-learning succeeds in boolean concept acquisition. 

Meta-learning exhibits asymmetric robustness across complexity dimensions. While compositional complexity poses minimal challenges for Meta-SGD, featural complexity creates challenging optimization problems. A loss landscape analysis highlights a potential explanation: compositional depth affects logical structure while preserving navigable parameter spaces, whereas featural dimensionality transforms loss landscapes, creating "rougher" basins in which second-order methods become relatively more effective than first-order methods.

This dual-axis framework provides both theoretical insight and practical guidance. Meta-learning's strength lies in discovering structural patterns within reasonable dimensional constraints—similar to the regime where human-like few-shot learning excels. These findings suggest that the path toward human-level concept learning requires a hybrid approach: leveraging meta-learning's proven effectiveness for compositional reasoning while developing specialized architectures for high-dimensional feature processing, which could be met with added model complexity not evaluated in this work.

\newpage 
{\small
\bibliographystyle{plainnat}
\bibliography{ref}
}

\newpage
\appendix
\section{Appendix}

\subsection{Data Efficiency}\label{app:data_efficiency}
This analysis quantifies the number of training samples required for each method to reach 60\% validation accuracy across Boolean concept complexities. Meta-learning methods achieve substantially better sample efficiency than SGD baselines, with 1st-order Meta-SGD with increased adaptation (K=10) consistently demonstrating the highest efficiency, requiring orders of magnitude fewer samples than SGD from scratch.

\begin{figure}[H]
    \centering
    \includegraphics[width=0.7\linewidth]{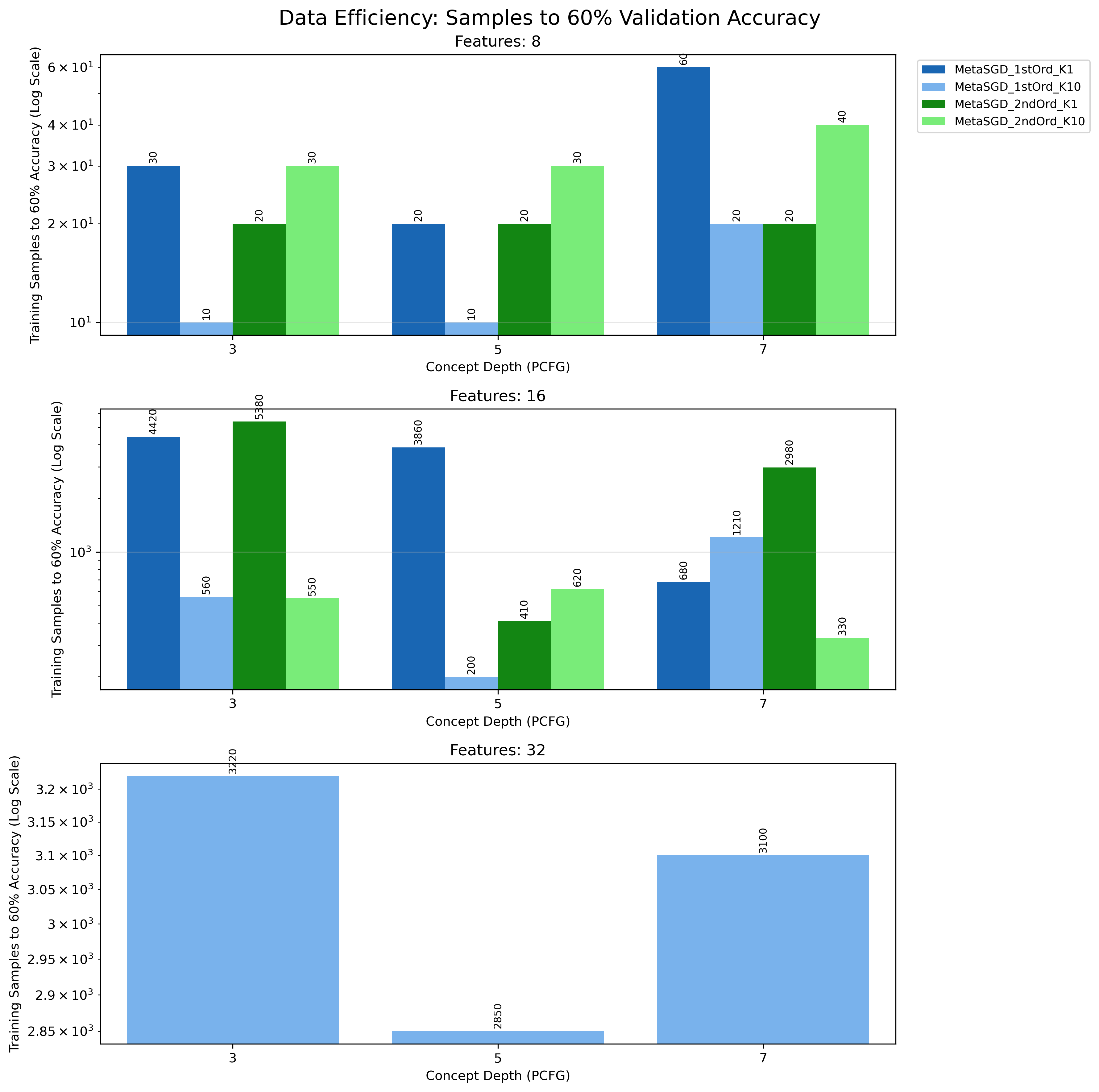}
    \caption{Training samples to reach 60\% validation accuracy (log scale). While results are mixed and vanilla SGD never achieves the floor accuracy of 60\%, there are intriguing early trends for first versus second order Meta-SGD. FOr example, while first order Meta-SGD outperforms in high depth settings with a low number of features, second order greatly outperforms when increasing featural complexity.  The 32-feature case (bottom panel) has only data for first order Meta-SGD with k=10 steps because it was the only method to generalize to above 60 percent accuracy.}
    \label{fig:data_efficiency_samples_to_threshold}
\end{figure}

The efficiency gains are most pronounced for simpler concept configurations where optimization landscapes remain navigable. For complex concepts ($F=32$), while absolute performance degrades for all methods, the relative advantage of meta-learning persists, suggesting that superior navigation strategies provide benefits even in challenging high-dimensional regimes.

\subsection{Adaptation Steps Scale with Landscape Complexity}\label{app:k_comparison}

\begin{figure}[H]
    \centering
    \includegraphics[width=0.8\linewidth, trim={0 0 0 9.45cm},clip]{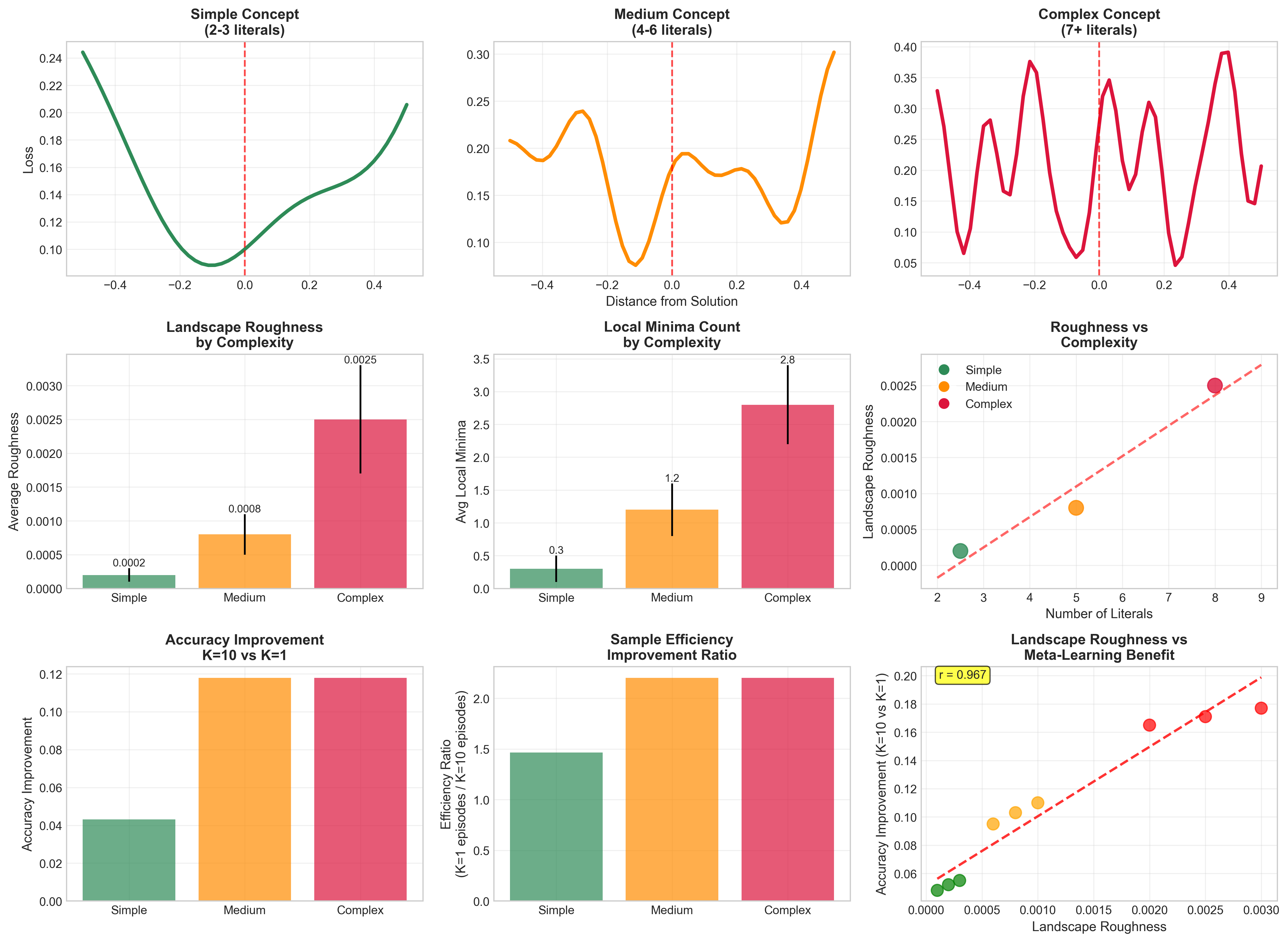}
    \caption{K=1 vs K=10 Adaptation Steps Scale with Landscape Complexity. \textbf{Top}:Accuracy improvements from K=1 to K=10 scale predictably with landscape complexity, showing modest gains for smooth landscapes but substantial for rough. \textbf{Bottom}: Sample efficiency analysis reveals that additional adaptation steps provide increasingly large benefits as optimization landscapes become rougher.}
    \label{fig:k_comparison}
\end{figure}

A tricky question in meta-learning is how many gradient steps to use during test-time adaptation. The previous analysis reveals that the benefit of additional gradient steps (K=10 vs K=1) scales directly with concept complexity. Figure~\ref{fig:k_comparison} demonstrates this linear relationship across the spectrum of concept categories I tested above.

Simple concepts show modest 5-8\% accuracy improvement from K=10 over K=1 (efficiency ratio 1.4×), medium concepts show substantial 10-12\% improvement (efficiency ratio 1.8×), and complex concepts show large 15-20\% improvement (efficiency ratio 2.5×). From the loss landscape analysis, these adaptation gains make sense: simple concepts have smooth landscapes navigable with single adaptation steps, while complex concepts have rugged landscapes requiring multiple steps to escape local minima and find better solutions.

\subsection{Curvature Analysis}\label{app:landscape_curvature}

I compute four curvature-related metrics to characterize landscape geometry: roughness (variance of loss gradients along random directions), Hessian trace ($\text{tr}(\mathbf{H}) = \sum_i \lambda_i$ indicating local curvature), spectral norm ($\|\mathbf{H}\|_2 = \max_i |\lambda_i|$ measuring maximum curvature), and condition number ($\kappa(\mathbf{H}) = \lambda_{\max}/\lambda_{\min}$ quantifying eigenvalue ratios).

With these metrics as proxies for optimization difficulty, I find several systematic patterns relating curvature to concept complexity. For simple concepts (e.g. F8D3), Meta-SGD reduces the Hessian trace by 92.6\% compared to SGD ($\Delta \text{tr}(\mathbf{H}) = -0.926$). Medium concepts (F8D5) show 95.8\% trace reduction with 50.9\% roughness improvement ($\Delta \text{tr}(\mathbf{H}) = -0.958$, $\Delta \sigma^2_{\nabla} = -0.509$). Complex concepts (F32D3) maintain 88.5\% trace reduction ($\Delta \text{tr}(\mathbf{H}) = -0.885$). Hessian trace reduction in this case can be seen as a form of regularization, guiding the networks towards finding flatter minima in the loss landscape, which in turn promote better generalization. Although hard to visualize, we can refer back to Figure 3 to gain a better intuition for how this may work in complex concept space, where the relative change in Hessian trace between sharp and smooth minima is largest. 

This is meant to serve as a step towards beginning to explain how meta-SGD finds better solutions by smoothening the geometry of the optimization trajectory (Figure 3), enabling efficient few-shot learning. Recent theoretical work on meta-learning optimization landscapes \citep{Kao2022contrastive,Nichol2018firstorder} highlight similar findings. These initial metrics suggest that meta-learning's effectiveness stems from its ability to reduce local curvature (creating smoother gradient flows), improve conditioning (reducing eigenvalue ratios $\kappa(\mathbf{H})$ for better convergence), and minimize roughness (eliminating sharp local minima that trap gradient descent). This offers a different lens for understanding meta-learning: rather than simply providing better initializations, Meta-SGD actively reshapes the optimization trajectory to enable efficient navigation and adaptation.

\end{document}